\documentclass[10pt,twocolumn,letterpaper]{article}

\usepackage[pagenumbers]{iccv} % To force page numbers, e.g. for an arXiv version

\usepackage{amsfonts}
\usepackage{amsmath}
\usepackage{amssymb}

\usepackage{makecell}
\usepackage{multicol}
\usepackage{multirow}
\usepackage{pifont}
\newcommand{\cmark}{\ding{51}}
\newcommand{\xmark}{\ding{55}}

\usepackage[table,dvipsnames]{xcolor}

\definecolor{t2e_red}{RGB}{239,99,75}
\definecolor{t2e_blue}{RGB}{99,113,250}
\definecolor{t2e_green}{RGB}{0,180,139}
\definecolor{t2e_gray}{RGB}{165,165,165}
\definecolor{redlink}{RGB}{239,99,75}

\usepackage[pagebackref=false,breaklinks,colorlinks,allcolors=redlink]{hyperref}

\hypersetup{
  colorlinks=true,
  linkcolor=redlink,
  citecolor=redlink,
  urlcolor=redlink,
}

\usepackage{listings}
\usepackage{tcolorbox}
\usepackage{framed}
\usepackage{fontawesome}

\newcommand\blfootnote[1]{%
\begingroup
\renewcommand\thefootnote{}{}\footnote{#1}%
\addtocounter{footnote}{-1}%
\endgroup
}

\title{
Visual Grounding from Event Cameras
}

\author{
Lingdong Kong$^{1,2,*}$
\quad
Dongyue Lu$^{1,*}$
\quad
Ao Liang$^{1,*}$
\quad
Rong Li$^{3}$
\quad
Yuhao Dong$^{4}$
\quad
Tianshuai Hu$^{5}$
\\
Lai Xing Ng$^{6}$
\quad
Wei Tsang Ooi$^{1}$
\quad
Benoit R. Cottereau$^{7,8}$
\\[0.5ex]
{\small$^1$NUS}
~~ 
{\small$^2$CNRS@CREATE}
\quad {\small$^3$HKUST(GZ)}
~~
{\small$^4$NTU}
~~
{\small$^5$HKUST}
~~
{\small$^6$I$^2$R, A*STAR}
~~
{\small$^7$IPAL, CNRS}
~~
{\small$^8$CerCo, CNRS}
\\[0.9ex]
\faGlobe~\textbf{Project Page:} \href{https://talk2event.github.io}{\textcolor{t2e_red}{\textbf{\textsl{Link}}}} 
~\quad~
\faGithubAlt~\textbf{GitHub:} \href{https://github.com/talk2event/toolkit}{\textcolor{t2e_green}{\textbf{\textsl{Link}}}}
~\quad~
\faDatabase~\textbf{Dataset:} \href{https://huggingface.co/datasets/talk2event/dataset}{\textcolor{t2e_blue}{\textbf{\textsl{Link}}}}
}

\def\paperID{NeVi 2025}
\def\confName{ICCV}
\def\confYear{2025}

\begin{document}

% \maketitle
\twocolumn[{
    \renewcommand\twocolumn[1][]{#1}
    \maketitle
    \begin{center}
    \vspace{-0.55cm}
    \includegraphics[width=\textwidth]{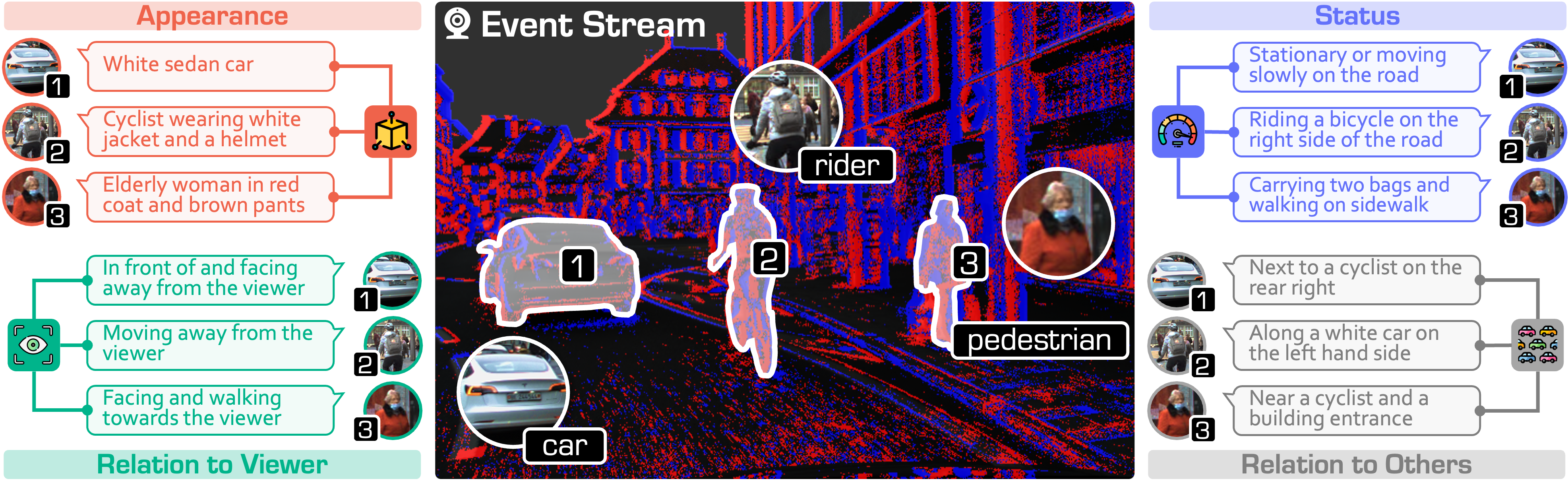}
    \vspace{-0.65cm}
    \captionof{figure}{\textbf{Grounded scene understanding from event cameras.} This work introduces \includegraphics[width=0.024\linewidth]{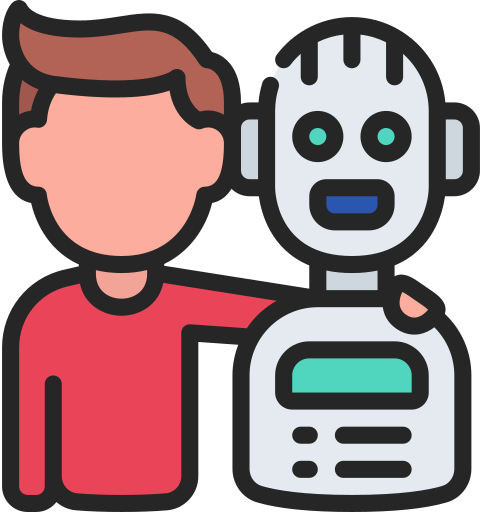}~\textsf{Talk2Event}, a novel task and dataset for localizing dynamic objects from event streams using natural language descriptions, where each unique object in the scene is defined by \textbf{four key attributes}: \textcolor{t2e_red}{\ding{172}}\textit{Appearance}, \textcolor{t2e_blue}{\ding{173}}\textit{Status}, \textcolor{t2e_green}{\textbf{\ding{174}}}\textit{Relation-to-Viewer}, and \textcolor{gray}{\textbf{\ding{175}}}\textit{Relation-to-Others}. We find that modeling these attributes enables precise, interpretable, and temporally-aware grounding across diverse dynamic environments in the real world.}
    \label{fig:teaser}
    \end{center}
}]

% Footnote
\blfootnote{$(*)$ Lingdong, Dongyue, and Ao contributed equally to this work.}
 
% Main Body
\begin{abstract}
Event cameras capture changes in brightness with microsecond precision and remain reliable under motion blur and challenging illumination, offering clear advantages for modeling highly dynamic scenes. Yet, their integration with natural language understanding has received little attention, leaving a gap in multimodal perception. To address this, we introduce \textbf{Talk2Event}, the first large-scale benchmark for \textit{language-driven object grounding} using event data. Built on real-world driving scenarios, Talk2Event comprises 5{,}567 scenes, 13{,}458 annotated objects, and more than 30{,}000 carefully validated referring expressions. Each expression is enriched with four structured attributes -- \textit{appearance}, \textit{status}, \textit{relation to the viewer}, and \textit{relation to surrounding objects} -- that explicitly capture spatial, temporal, and relational cues. This attribute-centric design supports interpretable and compositional grounding, enabling analysis that moves beyond simple object recognition to contextual reasoning in dynamic environments. We envision Talk2Event as a foundation for advancing multimodal and temporally-aware perception, with applications spanning robotics, human-AI interaction, and so on.
\end{abstract} 
\vspace{-0.15cm}
\section{Introduction}
\label{sec:intro}

Event-based sensors \cite{chakravarthi2024survey,gallego2022survey,steffen2019neuromorphic} are increasingly recognized as a compelling alternative to conventional frame cameras. Unlike standard sensors (\eg, RGB cameras), event cameras record brightness changes asynchronously with microsecond precision \cite{xu2023deep,brandli2014davis}, consume little power \cite{posch2010gen1,son2017gen3,finateu2020gen4}, and remain robust under motion blur and poor illumination \cite{rebecq2019e2vid,jeong2024towards,kamal2024efficient,ercan2024hue}. These advantages have enabled progress across diverse perception tasks, including object detection \cite{gehrig2023rvt,gehrig2024dagr,lu2024flexevent,lu2024flexevent}, semantic segmentation \cite{sun2022ess,hamaguchi2023hmnet,hareb2024evsegsnn,kong2025eventfly}, and visual odometry or SLAM \cite{censi2014odometry,mueggler2018continuous,hidalgo2022event}. Despite these advances, most works focus on geometric or low-level semantics, leaving one essential ability unexplored in the event domain: \textbf{visual grounding}, \ie, localizing objects from natural language descriptions.

\begin{table*}[t]
\centering
\caption{\textbf{Summary of benchmarks}. We compare datasets from aspects including: $^1$\textbf{Sensor} ({\includegraphics[width=0.024\linewidth]{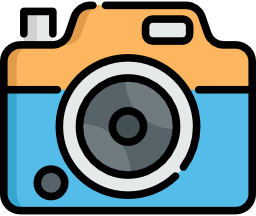}} Frame, {\includegraphics[width=0.024\linewidth]{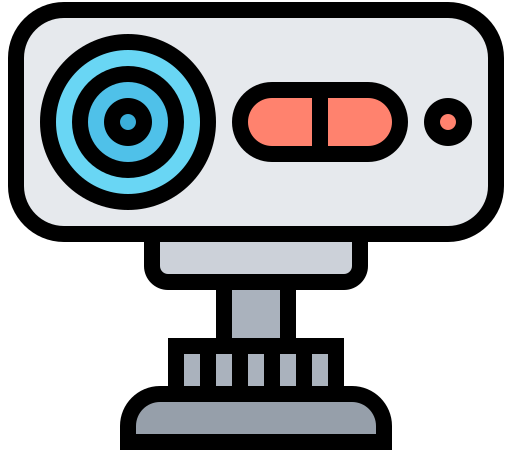}} RGB-D, {\includegraphics[width=0.025\linewidth]{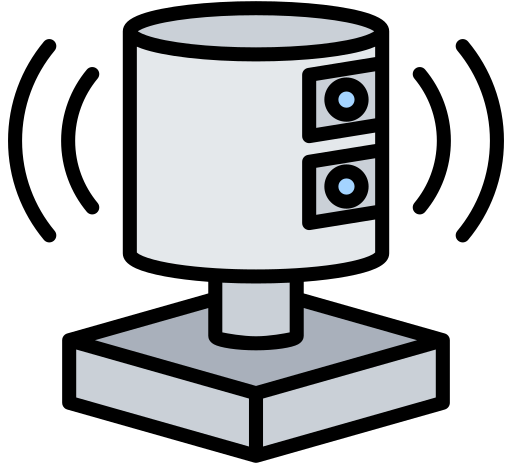}} LiDAR, {\includegraphics[width=0.025\linewidth]{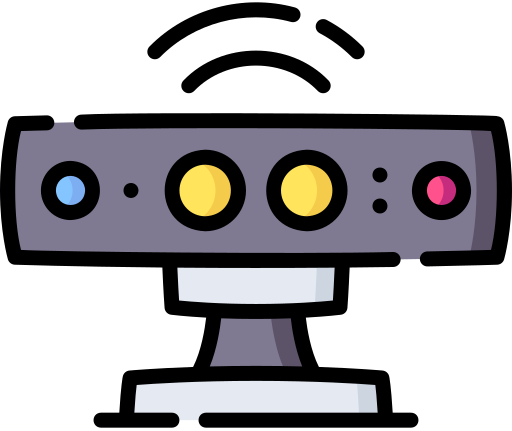}} Event), $^2$\textbf{Type}, $^3$\textbf{Statistics} (number of scenes, objects, referring expressions, and average length per caption), and supported $^4$\textbf{Attributes} for grounding, \ie, \textcolor{t2e_red}{\ding{172}}\texttt{Appearance} ($\delta_{\textcolor{t2e_red}{\mathrm{\mathbf{a}}}}$), \textcolor{t2e_blue}{\ding{173}}\texttt{Status} ($\delta_{\textcolor{t2e_blue}{\mathrm{\mathbf{s}}}}$), \textcolor{t2e_green}{\ding{174}}\texttt{Relation-to-Viewer} ($\delta_{\textcolor{t2e_green}{\mathrm{\mathbf{v}}}}$), \textcolor{darkgray}{\ding{175}}\texttt{Relation-to-Others} ($\delta_{\textcolor{darkgray}{\mathrm{\mathbf{o}}}}$).}
\vspace{-0.2cm}
\resizebox{0.9\linewidth}{!}{
\begin{tabular}{r|r|c|c|cccc|cccc}
    \toprule
    \multirow{2}{*}{\textbf{Dataset}} & \multirow{2}{*}{\textbf{Venue}} & \textbf{Sensory} & \textbf{Scene} & \multicolumn{4}{c|}{\textbf{Statistics}} & \multicolumn{4}{c}{\textbf{Attributes}}
    \\
    & & \textbf{Data} & \textbf{Type} & \textbf{Scene} & \textbf{Obj.} & \textbf{Expr.} & \textbf{Len.} & $\delta_{\textcolor{t2e_red}{\mathrm{\mathbf{a}}}}$ & $\delta_{\textcolor{t2e_blue}{\mathrm{\mathbf{s}}}}$ & $\delta_{\textcolor{t2e_green}{\mathrm{\mathbf{v}}}}$ & $\delta_{\textcolor{t2e_gray}{\mathrm{\mathbf{o}}}}$
    \\\midrule\midrule
    RefCOCO+ \cite{yu2016refcoco} & \textcolor{gray}{{\small ECCV'16}} & \raisebox{-0.2\height}{\includegraphics[width=0.024\linewidth]{figures/icons/frame.png}} & Static & $19{,}992$ & $49{,}856$ & $141{,}564$ & $3.53$ & \textcolor{t2e_green}{\cmark} & \textcolor{t2e_red}{\xmark} & \textcolor{t2e_red}{\xmark} & \textcolor{t2e_red}{\xmark} 
    \\
    RefCOCOg \cite{yu2016refcoco} & \textcolor{gray}{{\small ECCV'16}} & \raisebox{-0.2\height}{\includegraphics[width=0.024\linewidth]{figures/icons/frame.png}} & Static & $26{,}711$ & $54{,}822$ & $85{,}474$ & $8.43$ & \textcolor{t2e_green}{\cmark} & \textcolor{t2e_red}{\xmark} & \textcolor{t2e_red}{\xmark} & \textcolor{t2e_green}{\cmark} 
    \\
    Nr3D \cite{achlioptas2020referit3d} & \textcolor{gray}{{\small ECCV'20}} & \raisebox{-0.2\height}{\includegraphics[width=0.024\linewidth]{figures/icons/frame.png}} \raisebox{-0.2\height}{\includegraphics[width=0.024\linewidth]{figures/icons/rgbd.png}} & Static & $707$ & $5{,}878$ & $41{,}503$ & - & \textcolor{t2e_green}{\cmark} & \textcolor{t2e_red}{\xmark} & \textcolor{t2e_red}{\xmark} & \textcolor{t2e_green}{\cmark} 
    \\
    Sr3D \cite{achlioptas2020referit3d} & \textcolor{gray}{{\small ECCV'20}} & \raisebox{-0.2\height}{\includegraphics[width=0.024\linewidth]{figures/icons/frame.png}} \raisebox{-0.2\height}{\includegraphics[width=0.024\linewidth]{figures/icons/rgbd.png}} & Static & $1{,}273$ & $8{,}863$ & $83{,}572$ & - & \textcolor{t2e_green}{\cmark} & \textcolor{t2e_red}{\xmark} & \textcolor{t2e_red}{\xmark} & \textcolor{t2e_green}{\cmark}
    \\
    ScanRefer \cite{chen2020scanrefer} & \textcolor{gray}{{\small ECCV'20}} & \raisebox{-0.2\height}{\includegraphics[width=0.024\linewidth]{figures/icons/frame.png}} \raisebox{-0.2\height}{\includegraphics[width=0.024\linewidth]{figures/icons/rgbd.png}} & Static & $800$ & $11{,}046$ & $51{,}583$ & $20.3$ & \textcolor{t2e_green}{\cmark} & \textcolor{t2e_red}{\xmark} & \textcolor{t2e_red}{\xmark} & \textcolor{t2e_green}{\cmark}
    \\
    Text2Pos \cite{kolmet2022kitti360pose} & \textcolor{gray}{{\small CVPR'22}} & \raisebox{-0.25\height}{\includegraphics[width=0.025\linewidth]{figures/icons/lidar.png}} & Static & - & $6{,}800$ & $43{,}381$ & - & \textcolor{t2e_green}{\cmark} & \textcolor{t2e_red}{\xmark} & \textcolor{t2e_green}{\cmark} & \textcolor{t2e_red}{\xmark} 
    \\
    CityRefer \cite{miyanishi2023cityrefer} & \textcolor{gray}{{\small NeurIPS'23}} & \raisebox{-0.25\height}{\includegraphics[width=0.025\linewidth]{figures/icons/lidar.png}} & Static & - & $5{,}866$ & $35{,}196$ & - & \textcolor{t2e_green}{\cmark} & \textcolor{t2e_red}{\xmark} & \textcolor{t2e_red}{\xmark} & \textcolor{t2e_green}{\cmark}
    \\
    Ref-KITTI \cite{wu2024ref-kitti} & \textcolor{gray}{{\small CVPR'23}} & \raisebox{-0.2\height}{\includegraphics[width=0.024\linewidth]{figures/icons/frame.png}} & Static & $6{,}650$ & - & $818$ & - & \textcolor{t2e_green}{\cmark} & \textcolor{t2e_red}{\xmark} & \textcolor{t2e_green}{\cmark} & \textcolor{t2e_red}{\xmark}  
    \\
    M3DRefer \cite{zhan2024mono3dvg} & \textcolor{gray}{{\small AAAI'24}} & \raisebox{-0.2\height}{\includegraphics[width=0.024\linewidth]{figures/icons/frame.png}} & Static & $2{,}025$ & $8{,}228$ & $41{,}140$ & $53.2$ & \textcolor{t2e_green}{\cmark} & \textcolor{t2e_red}{\xmark} & \textcolor{t2e_green}{\cmark} & \textcolor{t2e_red}{\xmark}  
    \\
    STRefer \cite{lin2024wildrefer} & \textcolor{gray}{{\small ECCV'24}} & \raisebox{-0.2\height}{\includegraphics[width=0.024\linewidth]{figures/icons/frame.png}} \raisebox{-0.25\height}{\includegraphics[width=0.025\linewidth]{figures/icons/lidar.png}} & Static & $662$ & $3{,}581$ & $5{,}458$ & - & \textcolor{t2e_green}{\cmark} & \textcolor{t2e_red}{\xmark} & \textcolor{t2e_red}{\xmark} & \textcolor{t2e_red}{\xmark}
    \\
    LifeRefer \cite{lin2024wildrefer} & \textcolor{gray}{{\small ECCV'24}} & \raisebox{-0.2\height}{\includegraphics[width=0.024\linewidth]{figures/icons/frame.png}} \raisebox{-0.25\height}{\includegraphics[width=0.025\linewidth]{figures/icons/lidar.png}} & Static & $3{,}172$ & $11{,}864$ & $25{,}380$ & - & \textcolor{t2e_green}{\cmark} & \textcolor{t2e_red}{\xmark} & \textcolor{t2e_red}{\xmark} & \textcolor{t2e_red}{\xmark}
    % \\
    % WaterVG \cite{guan2025watervg} & \textcolor{gray}{{\small TITS'25}} & Frame \& Radar & - & $11,568$ & $34,987$ & -
    \\\midrule
    \raisebox{-0.2\height}{\includegraphics[width=0.025\linewidth]{figures/icons/friends.png}}
    \textsf{Talk2Event} & \textbf{Ours} & \raisebox{-0.2\height}{\includegraphics[width=0.025\linewidth]{figures/icons/frame.png}} \raisebox{-0.2\height}{\includegraphics[width=0.024\linewidth]{figures/icons/event-camera.png}} \raisebox{-0.25\height}{\includegraphics[width=0.025\linewidth]{figures/icons/lidar.png}} & \textbf{Dynamic} & $\mathbf{5{,}567}$ & $\mathbf{13{,}458}$ & $\mathbf{30{,}690}$ & $\mathbf{34.1}$ & \textcolor{t2e_green}{\cmark} & \textcolor{t2e_green}{\cmark} & \textcolor{t2e_green}{\cmark} & \textcolor{t2e_green}{\cmark}
    \\
    \bottomrule
\end{tabular}}
\label{tab:benchmarks}
\end{table*}

Visual grounding~\cite{xiao2024survey,liu2024survey} has become a cornerstone of multimodal perception, enabling human-AI interaction, vision-language navigation, and open-vocabulary recognition~\cite{wu2024survey,kong2024openess}. Benchmarks have been established for 2D images \cite{yang2019fast,yang2019cross,yang2022tubedetr}, videos \cite{liu2021context}, and 3D environments \cite{yang2021sat,yuan2024visual,chen2020scanrefer,achlioptas2020referit3d,zhao20213dvg-transformer,li2025seeground}, while more recent work extends grounding to point clouds \cite{yuan2024visual} and remote sensing \cite{sun2022visual,zhan2023rsvg,kuckreja2024geochat,zhou2024geoground}. These benchmarks, however, all rely on dense sensing modalities such as RGB frames or depth, which degrade in fast motion, high dynamic range, or low-light settings. Event cameras naturally mitigate these issues, but have not been studied in the context of grounding. Bridging asynchronous sensing with free-form natural language remains a crucial gap.

\textbf{Dynamic visual perception} research highlights the potential of events in scenarios where conventional cameras struggle. Large-scale driving and indoor datasets \cite{chaney2023m3ed,binas2017ddd17,gehrig2021dsec,zhu2018mvsec,perot2020mpx1} as well as synthetic benchmarks \cite{kim2021n-imagenet,gehrig2019representation,cho2023label} have enabled tasks ranging from detection \cite{gehrig2022pushing,maqueda2018event,yang2023ecdp} to action recognition \cite{plou2024eventsleep,zhou2024exact}. Robustness studies further show resilience to noise and illumination shifts \cite{chen2024ecmd,zhu2024cear,wang2024eventvot}. Yet, these efforts remain limited to geometry, appearance, or motion classification, without linking events to linguistic queries. On the other hand, \textbf{multimodal grounding} has progressed rapidly with region-ranking \cite{wang2018learning,wang2019neighbourhood} and transformer-based approaches \cite{kamath2021mdetr,jain2022butd-detr} on static datasets, later extended to video \cite{liu2023exploring} and RGB-D scenes \cite{chen2020scanrefer}. Despite this breadth, no prior work addresses how to align the sparse, asynchronous representations of event data with natural language supervision.

We address this gap with \textbf{Talk2Event}, the first dataset for \textit{language-driven object grounding} in event-based perception. Built on real-world driving scenarios from the large-scale DSEC \cite{gehrig2021dsec}, our constructed Talk2Event dataset provides $5{,}567$ scenes, $13{,}458$ annotated objects, and $30{,}690$ validated referring expressions. Each description is further labeled with four explicit attributes: \textcolor{t2e_red}{\ding{172}}\textit{Appearance}, \textcolor{t2e_blue}{\ding{173}}\textit{Status}, \textcolor{t2e_green}{\ding{174}}\textit{Relation-to-Viewer}, and \textcolor{gray}{\ding{175}}\textit{Relation-to-Others}. These attributes capture complementary spatiotemporal and relational cues, enabling compositional reasoning that goes beyond category- or geometry-level annotations. As shown in Fig.~\ref{fig:teaser}, Talk2Event offers multi-caption supervision with attribute-level annotations, establishing a new platform for studying multimodal, temporally-grounded visual grounding, and is tailored for event-based vision research.

\begin{figure*}[th]
    \centering
    \includegraphics[width=\linewidth]{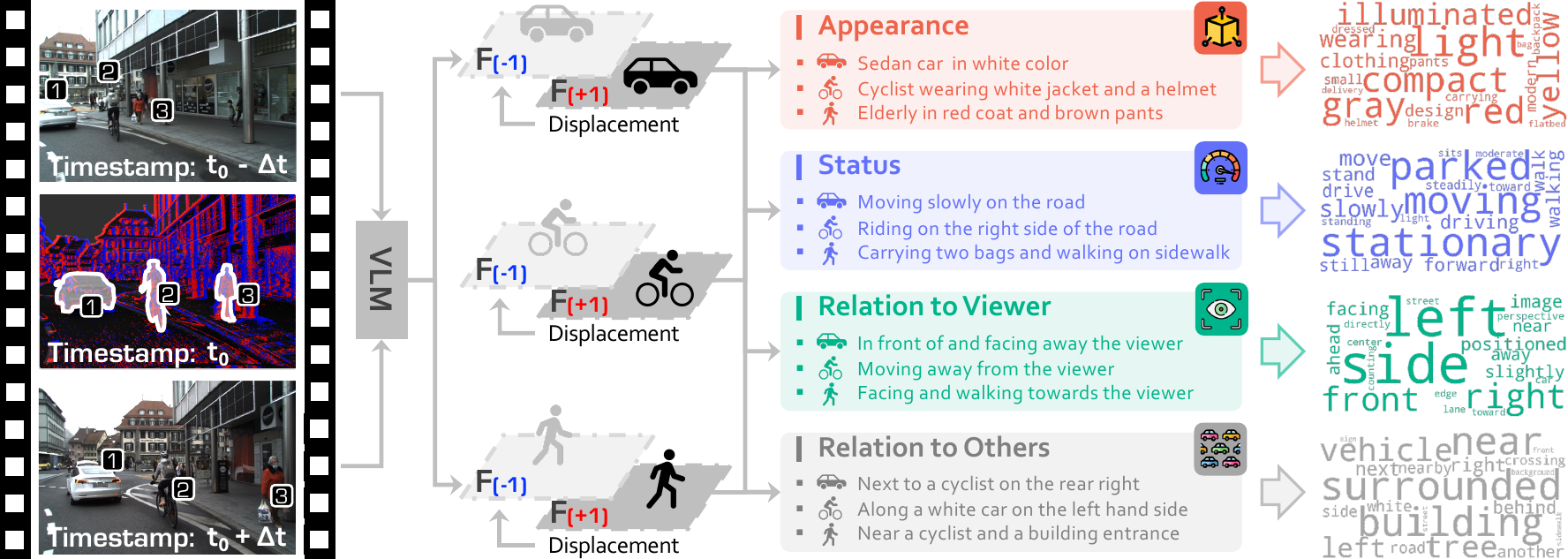}
    \vspace{-0.6cm}
    \caption{\textbf{Pipeline of dataset curation.} We leverage two surrounding frames at $t_0 \pm \Delta t$ to generate context-aware referring expressions of the event stream at $t_0$. Such a description covers key attributes: appearances, motion changes, spatial relations, and interactions. The word clouds shown on the right side highlight distinct linguistic patterns across the four grounding attributes.}
    \label{fig:pipeline}
\end{figure*}

In summary, the Talk2Event dataset aims to establish the first large-scale benchmark for event-based visual grounding. A key feature of the dataset is its attribute-centric annotation protocol, which encodes not only appearance and motion but also egocentric perspective and inter-object relations. This design enables interpretable, fine-grained, and compositional evaluation of grounding in dynamic environments, setting the stage for future research on multimodal and temporally-aware perception.
\section{Dataset \& Benchmark}
\label{sec:dataset}

We introduce \textbf{Talk2Event}, a benchmark designed to study language-driven grounding in event-based perception. This section first establishes the formal task definition and grounding objectives (Sec.~\ref{sec:task_formulation}), and then details the pipeline that transforms raw multimodal recordings into linguistically rich, attribute-aware annotations (Sec.~\ref{sec:dataset_curation}).

\subsection{Task Formulation}
\label{sec:task_formulation}

\noindent\textbf{Problem Definition.}  
Event-based visual grounding can be formulated as localizing an object within a dynamic scene captured by event cameras, conditioned on a free-form natural language description. Concretely, given a voxelized event representation $\mathbf{E}$ and a referring expression $\mathcal{S} = \{w_1, w_2, \dots, w_C\}$ consisting of $C$ tokens, the task is to output a bounding box $\hat{\mathbf{b}} = (x, y, w, h)$ that corresponds to the object described in $\mathcal{S}$.

Unlike conventional cameras that record intensity images at fixed intervals, event sensors produce an asynchronous stream $\mathcal{E} = \{e_k\}_{k=1}^N$, where each event $e_k = (x_k, y_k, t_k, p_k)$ specifies pixel location, timestamp, and polarity $p_k \in \{-1, +1\}$. To obtain a structured input compatible with modern backbones, we discretize this stream into a voxelized 4D tensor following~\cite{gehrig2023rvt,lu2024flexevent}, \ie:
\begin{equation}
\mathbf{E}(p, \tau, x, y) = \sum_{e_k \in \mathcal{E}} \delta(p - p_k) \, \delta(x - x_k, y - y_k) \, \delta(\tau - \tau_k),
\end{equation}
where $\tau_k = \left\lfloor \frac{t_k - t_a}{t_b - t_a} \times T \right\rfloor$ assigns the timestamp of event $e_k$ to one of $T$ temporal bins over the observation window $[t_a, t_b]$. The result, $\mathbf{E} \in \mathbb{R}^{2 \times T \times H \times W}$, retains spatiotemporal resolution and polarity, capturing the fine-grained dynamics of the scene.

\noindent\textbf{Benchmark Modalities.}  
Talk2Event does not restrict itself to events alone. Each sample is paired with a synchronized frame $\mathbf{F} \in \mathbb{R}^{3 \times H \times W}$ at reference time $t_0$. This enables three complementary evaluation settings: (i) using event voxels only, which emphasizes temporal dynamics; (ii) using the accompanying frame only, which emphasizes appearance cues; and (iii) combining both sources for multimodal grounding. Such a configuration allows researchers to study not only the strengths of each modality in isolation but also the benefits of cross-modal integration.

\noindent\textbf{Grounding Objectives.}  
To push beyond coarse or purely appearance-based grounding, each referring expression in Talk2Event is decomposed into four attribute categories that capture complementary cues:
\begin{itemize}
    \item \textcolor{t2e_red}{\textit{Appearance}}: static scene and object properties such as category, shape, size, or color. These cues align with traditional recognition tasks.  

    \item \textcolor{t2e_blue}{\textit{Status}}: dynamic aspects, \eg, whether the object is moving, stopped, turning, or crossing. These attributes leverage the temporal fidelity of events.  

    \item \textcolor{t2e_green}{\textit{Relation-to-Viewer}}: egocentric positioning relative to the observer, such as \emph{in front}, \emph{on the left}, \emph{far}, or \emph{facing the ego-vehicle}. This reflects view-conditioned grounding.  

    \item \textcolor{darkgray}{\textit{Relation-to-Others}}: contextual relations with surrounding objects, including both spatial layout (\eg, \emph{behind the bus}, \emph{next to the car}) and group behavior (\eg, \emph{two cyclists riding together}).  
\end{itemize}

\begin{figure*}[th]
    \centering
    \includegraphics[width=\linewidth]{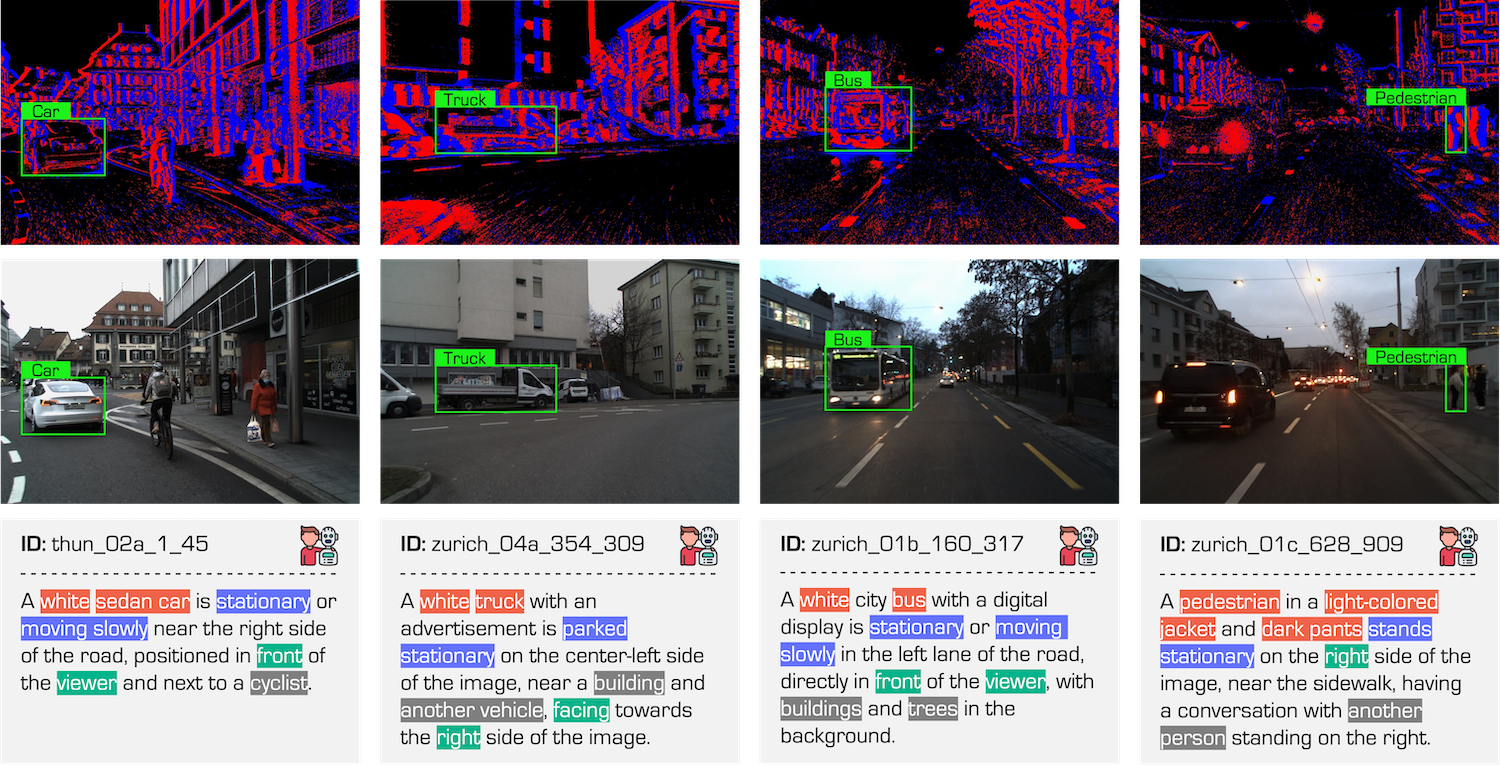}
    \vspace{-0.6cm}
    \caption{\textbf{Dataset examples.} We provide several event-based visual grounding examples from the \textbf{Talk2Event} dataset, spanning ``car'', ``truck'', ``bus'', and ``pedestrian'' classes. For more examples and semantic categories, kindly refer to the dataset page.}
    \label{fig:example}
\end{figure*}

By explicitly encoding these four dimensions, our dataset supports fine-grained, interpretable, and compositional evaluation. As highlighted in Tab.~\ref{tab:benchmarks}, prior benchmarks in images, video, or 3D rarely provide this structured supervision, and none exist for event-based data, leaving dynamic contexts underexplored.

\subsection{Data Curation Pipeline}
\label{sec:dataset_curation}

\noindent\textbf{Source Data.}  
Talk2Event is constructed from the DSEC dataset~\cite{gehrig2021dsec}, which offers synchronized event streams and high-resolution images across diverse driving environments. This multimodal foundation allows us to build the first benchmark that links event streams to natural language.

\noindent\textbf{Expression Generation.}  
To create linguistically rich descriptions, we design a context-aware prompting strategy (see Fig.~\ref{fig:pipeline}). For each object at time $t_0$, two neighboring frames at $t_0 - \Delta t$ and $t_0 + \Delta t$ ($\Delta t = 200$ ms) are provided to Qwen2-VL~\cite{qwen2-vl}. This temporal context encourages captions that mention not only static appearance but also displacement, motion, and relational cues. Each object is described by three distinct captions, which are subsequently refined through human verification for correctness and diversity. On average, descriptions contain 34.1 words, making Talk2Event significantly more verbose than existing grounding datasets. Attribute-specific word clouds (see Fig.~\ref{fig:pipeline}) further illustrate the coverage across appearance, motion, and relational cues.

\noindent\textbf{Attribute Decomposition.}  
Beyond raw captions, in our dataset, each referring expression is annotated with attribute labels for appearance, status, relation-to-viewer, and relation-to-others. This is achieved through a semi-automated pipeline: fuzzy matching and LLM-assisted parsing generate candidate labels, which are then verified by human annotators. This two-stage process ensures both scalability and semantic accuracy, while providing interpretable supervision for future models.

\noindent\textbf{Quality Assurance.}  
To guarantee the benchmark reliability, we adopt several filtering stages: (i) visibility checks discard heavily occluded, tiny, or ambiguous objects; (ii) redundancy checks enforce diversity by eliminating duplicate or near-identical captions; and (iii) attribute validation ensures that each caption references at least one meaningful attribute. After filtering, the Talk2Event dataset contains $5{,}567$ curated scenes, $13{,}458$ annotated objects, and $30{,}690$ validated referring expressions.

\noindent\textbf{Discussion.}  
Talk2Event converts raw event streams and frames into a benchmark with linguistically expressive and attribute-aware annotations. Unlike prior grounding datasets built on static frames or depth, it leverages asynchronous events with synchronized images to capture both temporal dynamics and appearance cues. Each object is described by multiple validated captions, ensuring correctness and diversity, as shown in \cref{fig:example}. The dataset’s structured attributes which cover appearance, motion, egocentric relations, and inter-object context reflect the complexity of real driving scenarios.  
This design enables systematic studies of modality-specific strengths, multimodal fusion, and compositional reasoning under dynamic conditions. It also supports robustness analysis in settings such as motion blur and low light, where frame-based benchmarks fail. By filling this gap, Talk2Event provides a unique foundation for advancing multimodal and temporally grounded perception in vision, language, and robotics.

\section{Conclusion}
\label{sec:conclusion}

We introduced \textbf{Talk2Event}, the first benchmark dedicated to language-driven grounding in event-based perception. Built on real-world driving data, the dataset provides $5{,}567$ curated scenes, $13{,}458$ annotated objects, and $30{,}690$ validated referring expressions, each enriched with four attribute categories that capture appearance, motion, egocentric relations, and inter-object context. Through a careful pipeline combining multimodal prompting, attribute decomposition, and human verification, our dataset delivers linguistically rich and interpretable annotations that expose the unique challenges of grounding in dynamic environments. We believe this resource will serve as a foundation for advancing research on multimodal and temporally-aware perception for event-based vision research.

% References
\clearpage
\clearpage
{
    \small
    \bibliographystyle{ieeenat_fullname}
    \bibliography{main}
}

\end{document}